\documentclass[journal]{IEEEtran}

\ifCLASSINFOpdf
\else
   \usepackage[dvips]{graphicx}
\fi
\usepackage{url}

\usepackage{graphicx}

\usepackage{xcolor}

\usepackage{tabularx}

\usepackage{arydshln}

\newcommand{\orcid}[1]{\href{https://orcid.org/#1}{\textcolor[HTML]{A6CE39}{\aiOrcid}}}

\usepackage[colorlinks=true, pdfstartview=FitV, linkcolor=blue, citecolor=blue, urlcolor=blue]{hyperref}

\usepackage{orcidlink}

\begin{document}

\title{A TextGCN-Based Decoding Approach for Improving Remote Sensing Image Captioning}

\author{Swadhin Das\orcidlink{0009-0004-1247-3275},~\IEEEmembership{Student Member, IEEE}, Raksha Sharma\,\orcidlink{0000-0003-2905-0194},~\IEEEmembership{Member, IEEE}
\thanks{Both the authors are from Department of Computer Science and Engineering, Indian Institute of Technology, Roorkee, India (e-mails are: \{s\_das,raksha.sharma\}@cs.iitr.ac.in).}
}

\maketitle

\begin{abstract}
Remote sensing images are highly valued for their ability to address complex real-world issues such as risk management, security, and meteorology. However, manually captioning these images is challenging and requires specialized knowledge across various domains. This letter presents an approach for automatically describing (captioning) remote sensing images. We propose a novel encoder-decoder setup that deploys a Text Graph Convolutional Network (TextGCN) and multi-layer LSTMs. The embeddings generated by TextGCN enhance the decoder's understanding by capturing the semantic relationships among words at both the sentence and corpus levels. Furthermore, we advance our approach with a comparison-based beam search method to ensure fairness in the search strategy for generating the final caption. We present an extensive evaluation of our approach against various other state-of-the-art encoder-decoder frameworks. We evaluated our method using seven metrics: BLEU-1 to BLEU-4, METEOR, ROUGE-L, and CIDEr. The results demonstrate that our approach significantly outperforms other state-of-the-art encoder-decoder methods.
\end{abstract}

\IEEEpeerreviewmaketitle
\section{Introduction}
Image captioning has been widely studied for natural images~\cite{vinyals2015show, karpathy2015deep}. However, remote sensing (RS) images are significantly different and more complex in their attributes compared to natural images~\cite{hoxha2020toward}. Recently, the captioning of RS images has gained the attention of artificial intelligence researchers. Initial attempts to address this problem utilized deep neural models for high-resolution images. Yuan et al.~[2019]~\cite{yuan2019exploring} proposed an image captioning framework based on multi-level attention and multi-label attribute graph convolution. Huang et al.~[2020]~\cite{huang2020denoising} introduced a denoising-based multi-scale feature fusion (DMSFF) mechanism that aggregates multi-scale features with denoising during visual feature extraction, aiding the encoder-decoder framework in obtaining denoised multi-scale feature representations. Wang et al.~[2020]~\cite{wang2020retrieval} proposed a topic retrieval model for caption generation, where the corresponding five sentences for each training image are combined as a collection of words. Furthermore, Li et al.~[2021]~\cite{li2021recurrent} proposed a recurrent attention and semantic gate framework that integrates competitive visual features with a recurrent attention mechanism. Hoxha et al.~[2021]~\cite{hoxha2021novel} developed a network of Support Vector Machines (SVMs) for decoding image information into descriptions. Zhao et al.~[2021]~\cite{zhao2021high} developed a fine-grained, structured, attention-based method that leverages the semantic structure of image contents. Li et al.~[2021]~\cite{li2021recurrent} proposed a recurrent attention and semantic gate-based method. Zhang et al.~[2021]~\cite{zhang2021global} introduced a model guided by global visual features and linguistic state attention. Wang et al.~[2022]~\cite{wang2022multiscale} developed a multi-scale multi-interaction network (MMN). Ye et al.~[2022]~\cite{ye2022joint} proposed a joint-training two-stage captioning method using multi-label classification. Liu et al.~[2022]~\cite{liu2022remote} introduced a multi-layer aggregated transformer model for remote sensing captioning. Wang et al.~[2022]~\cite{wang2022glcm} proposed an attention-based global-local captioning model (GLCM) to obtain global-local visual feature representations for captioning. Ren et al.~[2022]~\cite{ren2022mask} introduced a mask-guided Transformer network with a topic token to enhance the accuracy and diversity of captions.

While these models have shown significant performance in remote sensing image captioning (RSIC), they remain prone to overfitting due to limitations in training data~\cite{shen2020remote}, leading to misclassification of many complex images and weakened generated captions. To mitigate this issue, we utilize a Text Graph Convolutional Network (TextGCN) embedding technique, which assigns a unique embedding vector to each word throughout the model. This approach enhances the model's understanding of contextual words and addresses the data scarcity issue in RS image captioning.

Another major limitation of these existing models is the insufficiency of search mechanisms in accurately predicting captions. To address this, Hoxha et al.(2020)~\cite{hoxha2020new} proposed a comparison-based beam search method. They compared all candidate captions generated in the final round of beam search with captions from similar images, selecting the best caption based on this comparison. However, their model relied solely on the BLEU-2 score, which focuses only on sentence precision. Our approach introduces two key upgrades to the comparison-based beam search to enhance the quality of generated captions. First, we used the arithmetic mean of BLEU-2, METEOR, and ROUGE scores to calculate the sentence score, thereby balancing precision (BLEU), recall (METEOR), and the longest common subsequences (ROUGE). Second, we included an additional sentence generated by greedy search (if it wasn’t already part of the beam search-generated captions) in the comparison, making the search strategy versatile. The primary contribution of this letter is as follows:
\begin{itemize}
\item We propose a novel encoder-decoder setup that deploys a Text Graph Convolutional Network (TextGCN) and multi-layer LSTMs. TextGCN enhances the understanding of the decoder by capturing complex semantic relationships among words in the corpus.
\item Furthermore, we enhance our approach with a novel comparison-based beam search method to ensure that the generated captions closely match the gold-standard captions.
\end{itemize}

The subsequent sections of this letter are organized as follows: Section~\ref{sec:proposed} describes the components and techniques of our model. Section~\ref{sec:experiment} presents the experimental setup and results. Finally, Section~\ref{sec:conclusions} concludes the letter.

\section{Proposed Method}
\label{sec:proposed}
\begin{figure}[!ht]
  \centering
  \includegraphics[width=0.475\textwidth]{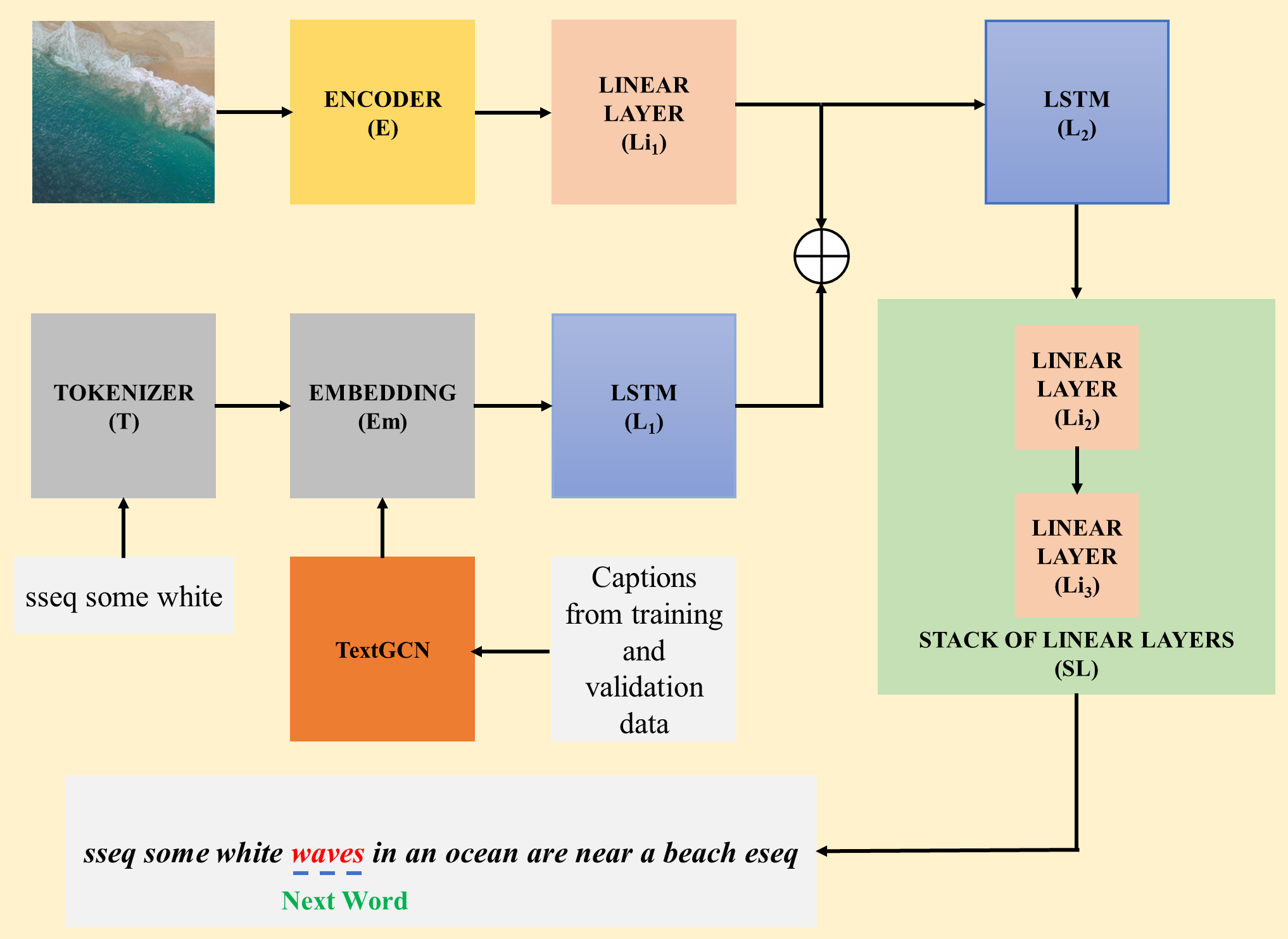}
   \caption{Architecture of the Proposed Model}
   \label{fig_architecture}
\end{figure}

We present an encoder-decoder architecture that implements a multi-layer decoding strategy to generate textual descriptions of remote sensing images. The encoder module, detailed in Section~\ref{image} and Section~\ref{sequence}, is responsible for encoding the input image and the input word sequence, respectively. During the decoding process, we employ two Long Short-Term Memory (LSTM) networks. The output from the first LSTM ($L_1$) is concatenated with the encoded image vector and then passed to the second LSTM ($L_2$). The output from $L_2$ is further processed through a stack of linear layers. The final linear layer ($Li_3$), a SoftMax layer, produces an output with a size equal to the vocabulary, providing probabilities for the occurrence of each word in the vocabulary. The overall architecture of the model is shown in Fig.~\ref{fig_architecture}.

\subsection{Encoded Representation of Image}
\label{image}
RESNET~\cite{he2016deep} is an image encoder that uses residual blocks to effectively mitigate the vanishing gradient problem, allowing for the training of very deep neural networks. Residual connections in ResNet help mitigate the vanishing gradient problem, enabling the training of much deeper networks. They allow gradients to flow directly through the network, improving model convergence. Since our goal is to extract image features rather than perform classification, we exclude the final layer of RESNET. Instead, we use the output of the second-to-last layer as the extracted features corresponding to the input image.

\subsection{Encoded Representation of the Input Text}
\label{sequence}
To encode the input sequence, we initialize a fixed-length array, padded with zeros, to match the length of the longest caption in the training/validation dataset. The input sequence is processed through an embedding layer, projecting it into an embedding space denoted as $\mathbf{E_s}$. The output of the embedding layer, with dimensions $\mathbf{l\times h}$ (where \textit{l} is the sequence length and \textit{h} represents the dimensions of the embedding space), is then sent to the decoding module. Importantly, the weights of the decoder's embedding layer are initialized by the embedding matrix obtained from TextGCN (refer to Section~\ref{TGCN}). The weights remain fixed throughout the decoding process.

\subsection{Text Graph Convolution Network}
\label{TGCN}
The embedding layer is essential in any encoder-decoder model, as it transforms each token (typically words) into a meaningful vector, which the decoder then uses to interpret the input sequence. The weights of the embedding layer can be either trainable or non-trainable. When pretrained embeddings are employed, the layer becomes non-trainable, and each token maintains a unique representation throughout the training process. Numerous pretrained embeddings are available, but the considered task is highly domain-specific, leading to many tokens being absent from the pretrained embeddings. TextGCN has been widely used to generate embeddings of words capturing the the relation among words in the corpus~\cite{li2023decoding,kim2024stad}. TextGCN is based on the concept of Graph Convolution Network (GCN). It generates embeddings of words considering the relation among words at sentence level as well as corpus level. This two level information strengths the decoder to interpret the role of a word in a caption. In our work, we have trained the Text Graph Convolution Network (TextGCN) on the remote sensing (training and validation) dataset to generate domain-specific embeddings of each token in the vocabulary. 

The formulation for obtaining the embedding matrix (embedding vectors for all tokens) using TextGCN~\cite{yao2019graph} is shown in Equation \ref{eqn_TextGCN}.
\begin{eqnarray}
F_1 &=& D^{-\frac{1}{2}}AD^{-\frac{1}{2}}F_0W_0 \nonumber \\
F_2 &=& D^{-\frac{1}{2}}AD^{-\frac{1}{2}}F_1W_1
\label{eqn_TextGCN}
\end{eqnarray}

Here, the adjacency matrix~\emph{A} is constructed based on the Point-wise Mutual Information (PMI) between two words, using a fixed-size window applied to each sentence. The degree matrix is denoted as~\emph{D} which is a diagonal matrix calculated by counting non zero elements in the corresponding row of the adjacency matrix.~\emph{$F_i$} represents the feature matrix and~\emph{$W^i$} denotes the weight matrix at the~\emph{$i^{th}$} layer. The feature matrix can be initialized in various ways, with one popular method being the initialization of a diagonal matrix where the diagonal elements are the sum of the corresponding rows. However, since our aim is to find the embedding matrix, we have applied some extensions. We have taken the ceiling value of the diagonals and then project it onto an embedding initializer (\emph{torch.nn.Embedding} in our work). The weight matrices $W_0$ and $W_1$ are initialized using the~\emph{Normalized Xavier}~\cite{glorot2010understanding} weight initialization. Both layers apply a sigmoid activation function and standard Gaussian normalization to each row. At the end of this process, each row is divided by its maximum absolute value. The resulting feature matrix values from TextGCN serve as pre-calculated weights for the embedding layer, which is non-trainable (refer to Figure~\ref{fig_architecture}).

\subsection{Comparison Based Beam Search}
\label{beam}
Traditional beam search operates based on the log likelihood of the output probabilities for the generated captions, which may not be optimal in all cases. In contrast, the comparison-based beam search~\cite{hoxha2020new} considers not only the likelihood but also the quality of the generated captions by comparing them with reference captions (captions of similar images). This method evaluates all generated captions at the end of the beam search, comparing them to the reference captions rather than further calculating the log likelihood of each candidate caption. This approach helps to select the most accurate caption both qualitatively and quantitatively. We have enhanced the fairness of the comparison-based beam search by incorporating three evaluation matrices. In Hoxha et al.~\cite{hoxha2020new}, the comparison was performed using only the~\emph{BLEU-2} score, we have expanded this to consider the arithmetic mean of~\emph{BLEU-2},~\emph{METEOR}, and~\emph{ROUGE-L} to balance precision (\emph{BLEU}), recall (\emph{METEOR}), and the longest common subsequences (\emph{ROUGE-L}). Additionally, to make the searching more versatile~\cite{das2024unveiling}, we have included the caption generated by greedy search if it is not a part of the search space.

\subsection{Multi-Layer Decoding Strategy}
\label{multi}
A conventional encoder-decoder model typically uses a single decoder that takes the embedding matrix of the text input sequence and produces an output text feature vector. This output text feature vector is then concatenated (or added) with the input image feature vector. A stack of linear layers processes it to predict the next word. We have extended this idea by incorporating two LSTMs at different stages (refer to Fig.~\ref{fig_architecture}). The concept adhered is that, instead of directly sending the concatenated output of the first LSTM and the image feature vectors to the stack of linear layers (SL), it is passed through another LSTM. This multi-layer decoding strategy~\cite{wang2021cascade} helps the model gain a better understanding of the image, as its output is influenced by both the image and corresponding text features. Finally, the SL layer handles the decision-making process.

\section{Experiments}
\label{sec:experiment}
\subsection{Dataset and Performance Metrics}
In this study, we used three publicly available remote sensing image dataset:~\emph{RSICD}~\cite{lu2017exploring},~\emph{UCM}~\cite{qu2016deep}, and~\emph{SYDNEY}~\cite{qu2016deep}. RSICD is a balanced dataset with a reasonably good number of instances in comparison to the other two dataset  \cite{lu2017exploring}. Hence, in this letter, we have presented results exclusively from the RSICD dataset. Our analysis of these dataset revealed spelling and grammatical errors, as well as diverse country-specific variations in certain words within the dataset. To ensure consistency, we have corrected these issues by standardizing to American English conventions. The updated dataset are also referenced in the supplementary file. Our captioning system is thoroughly evaluated using seven metrics: BLEU-1 to BLEU-4, METEOR, ROUGE-L, and CIDEr. The evaluation involves calculating BLEU, METEOR, and CIDEr with the~\emph{nlgeval}~\cite{sharma2017nlgeval} package, while the ROUGE-L metric is computed using standard Python modules.  

\subsection{Experimental Setup}
In this study, we have used the same train-validation-test split configuration as the source dataset. Our methodology employs RESNET as the image encoder, with the extracted image feature size set to $2048$. The TextGCN embedding size in our model is set to $256$. The first LSTM network ($L_1$) has a hidden layer size of 256, while the second LSTM network ($L_2$) has a hidden layer size of $512$. The linear layers $Li_1$, $Li_2$, and $Li_3$ have dimensions of 256, 512, and the vocabulary size, respectively. Their activations are GELU for $Li_1$ and $Li_2$, and SoftMax for $Li_3$. We have applied a Dropout of $0.5$ before~\emph{$Li_1$} and both LSTM layers to avoid overfitting. Optimization involves using the~\emph{Adam} optimizer and a categorical cross-entropy loss function, with the learning rate set to the default rate of the~\emph{Adam} optimizer. From an experimental standpoint, our model undergoes training for $64$ epochs with an early stopping criteria~\cite{genccay2001pricing} with patience value as five. In the comparison based beam search method, beam size and the maximum number of captions are set to five. Four similar images for the input image are selected from the archive. This archive is created using the training and validation images along with their captions, while results (Table \ref{exp_rsicd}) are reported on test images. Using Euclidean distance as the similarity metric, we have employed the K-nearest Neighbour (KNN) algorithm to identify the most analogous images for comparing input and reference images.

\subsection{Results and Analysis}
We report results for both the numerical (Table~\ref{exp_rsicd}, Table~\ref{embed_rsicd}) and visual findings (Figure~\ref{fig_visual}) from the experiments, along with the effects of varying size of embeddings (Table~\ref{hyper_rsicd}).
\subsubsection{The Performance of the Proposed Method}
\begin{table}[!ht]
\setlength{\tabcolsep}{0.01mm}
\begin{center}
\caption{Experimental Results for Captioning on RSICD Dataset}
\label{exp_rsicd}
\begin{tabular}{|c|c|c|c|c|c|c|c|}
\hline
Models & BLEU-1 & BLEU-2 & BLEU-3 & BLEU-4 & METEOR & ROUGE-L & CIDEr \\
\hline
RBOW~\cite{lu2017exploring} & 0.440 & 0.238 & 0.151 & 0.104 & 0.168 & 0.360 & 0.467 \\
LFV~\cite{lu2017exploring} & 0.434 & 0.245 & 0.163 & 0.117 & 0.171 & 0.382 & 0.653 \\
CSMLF~\cite{wang2019semantic} & 0.511 & 0.291 & 0.190 & 0.135 & 0.169 & 0.379 & 0.339 \\
MGRUD~\cite{hoxha2021novel} & 0.603 & 0.425 & 0.320 & 0.252 & 0.230 & 0.438 & 0.659 \\
SVMDC~\cite{hoxha2021novel} & 0.600 & 0.435 & 0.355 & 0.269 & 0.230 & 0.456 & 0.685 \\
\hdashline
E-N-B & 0.617 & 0.442 & 0.340 & 0.273 & 0.259 & 0.449 & 0.765 \\
E-T-B & 0.625 & 0.453 & 0.349 & 0.281 & 0.259 & 0.460 & 0.786 \\
M-N-B & 0.625 & 0.455 & 0.352 & 0.283 & 0.265 & 0.462 & 0.796 \\
M-T-B & 0.631 & 0.459 & 0.357 & 0.289 & 0.261 & 0.463 & 0.802 \\
\hdashline
E-N-C & 0.631 & 0.456 & 0.352 & 0.282 & 0.260 & 0.457 & 0.787 \\
E-T-C & 0.636 & 0.462 & 0.355 & 0.285 & 0.260 & 0.466 & 0.805 \\
M-N-C & 0.644 & 0.471 & 0.364 & 0.292 & 0.266 & 0.475 & 0.819 \\
\textcolor{red}{M-T-C} & \textbf{0.651} & \textbf{0.482} & \textbf{0.375} & \textbf{0.308} & \textbf{0.275} & \textbf{0.480} & \textbf{0.827} \\
\hline
\end{tabular}
\end{center}
\end{table}
 Table~\ref{exp_rsicd} compares various baseline models and search techniques on the RSICD dataset. In the first section of Table~\ref{exp_rsicd} , we present the results for 1. RNN with Bag of Words features representation (RBOW)~\cite{lu2017exploring}, 2. LSTM with Fisher Vector features representation (LFV)~\cite{lu2017exploring}, 3. Collective Semantic Metric Learning Framework (CSMLF)~\cite{wang2019semantic}, 4. Merged GRU Decoder model (MGRUD)~\cite{hoxha2021novel}, and 5. SVM based Decoder model (SVMDC)~\cite{hoxha2021novel}. The subsequent sections evaluate different search techniques. The naming convention used is as follows: model (\emph{E} for conventional encoder-decoder and~\emph{M} for the multi-layer decoder-based RSIC model used in our work) - embeddings (\emph{N} for no pretrained embeddings and~\emph{T} for TextGCN embeddings) - search method (~\emph{B} for traditional beam search, and~\emph{C} for the comparison-based beam search~\cite{hoxha2020new} after modifications). Table~\ref{exp_rsicd} demonstrates that the proposed method, which combines the multi-layer decoder, TextGCN embeddings, and comparison-based beam search (M-T-C), yields the best results. We observed that the multi-layer decoder performs better than the conventional encoder-decoder models, and the comparison-based beam search also improves the quality of the captions compared to traditional searching techniques.
\subsubsection{The Effect of Pretrained Word Embeddings}
\begin{table}[!ht]
\setlength{\tabcolsep}{0.5mm}
\begin{center}
\caption{Experimental Results from Different Embeddings on RSICD dataset}
\label{embed_rsicd}
\begin{tabular}{|c|c|c|c|c|c|c|c|}
\hline
Embed & BLEU-1 & BLEU-2 & BLEU-3 & BLEU-4 & METEOR & ROUGE-L & CIDEr \\
\hline
None & 0.618 & 0.444 & 0.338 & 0.268 & 0.253 & 0.456 & 0.736 \\
GL & 0.620 & 0.446 & 0.341 & 0.272 & 0.25 & 0.457 & 0.744 \\
WV & 0.623 & 0.449 & 0.344 & 0.274 & 0.253 & 0.456 & 0.765 \\
FT & 0.620 & 0.447 & 0.340 & 0.269 & 0.249 & 0.456 & 0.763 \\
TG & \textbf{0.633} & \textbf{0.461} & \textbf{0.358} & \textbf{0.288} & \textbf{0.255} & \textbf{0.469} & \textbf{0.794} \\
\hline
\end{tabular}
\end{center}
\end{table}
We examined the impact of using pretrained word embeddings in the embedding layer compared to a trainable embedding layer. Table \ref{embed_rsicd} provides a comparison of different word embeddings within an encoder-decoder framework~\cite{qu2016deep} on the RSICD dataset. Here,~\emph{Embed} in the heading refers to the name of the embedding. Additionally,~\emph{None}~\cite{qu2016deep} indicates that no pretrained embeddings are used, meaning the embedding layer is trainable;~\emph{GL}~\cite{pennington2014glove} denotes GloVe;~\emph{WV}~\cite{mikolov2013efficient} denotes Word2Vec;~\emph{FT}~\cite{bojanowski2017enriching} denotes FastText; and~\emph{TG}~\cite{yao2019graph} denotes TextGCN. This result underscores the importance of using pretrained embeddings, demonstrating that TextGCN used in our work outperforms both trainable embeddings and other pretrained embedding methods.
\subsubsection{Effect of Embedding Vector Size of TextGCN on Our Model}
\begin{table}[!ht]
\setlength{\tabcolsep}{0.4mm}
\begin{center}
\caption{Hyper-parameter Results for Different Embedding Sizes of TextGCN by our Approach on RSICD Dataset}
\label{hyper_rsicd}
\begin{tabular}{|c|c|c|c|c|c|c|c|}
\hline
Size & BLEU-1 & BLEU-2 & BLEU-3 & BLEU-4 & METEOR & ROUGE-L & CIDEr \\
\hline
64 & 0.632 & 0.462 & 0.361 & 0.290 & 0.256 & 0.464 & 0.790 \\
128 & 0.636 & 0.462 & 0.357 & 0.287 & 0.259 & 0.466 & 0.808 \\
256 & \textbf{0.651} & \textbf{0.482} & \textbf{0.375} & \textbf{0.308} & \textbf{0.275} & \textbf{0.480} & \textbf{0.827} \\
512 & 0.641 & 0.466 & 0.363 & 0.294 & 0.262 & 0.468 & 0.810 \\
\hline
\end{tabular}
\end{center}
\end{table}
To analyze the influence of TextGCN embedding vector size, we trained our model with five different sizes $\left\{64, 128, 256, 512\right\}$ to cover a range of dimensions and evaluate their impact on performance. Table \ref{hyper_rsicd} presents the results of the impact of these four vector sizes on our model using the RSICD dataset. We observed that the performance was optimal with an embedding size of $256$. Additionally, while larger vector sizes did not significantly improve model performance, they increased training time. Consequently, we selected a vector size of 256 for our experiments.
\subsubsection{Visual Examples}
\begin{figure}[!ht]
  \centering
  \includegraphics[width=0.475\textwidth]{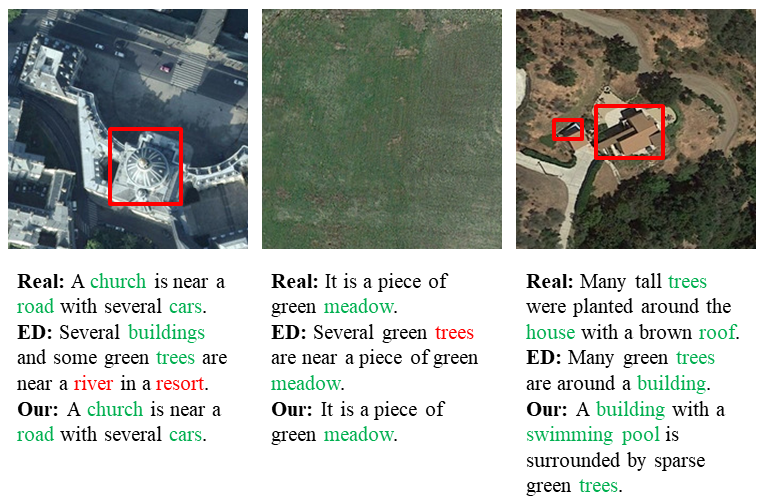}
   \caption{Visual Examples of Proposed RSIC Model}
   \label{fig_visual}
\end{figure}
Figure~\ref{fig_visual} presents some test images from the dataset along with their actual (Real) and predicted captions from both the encoder-decoder (ED) model with beam search (E-N-B) and our (Our) model (M-T-C). This comparison underscores the effectiveness of our model in accurately identifying key elements within the images. For the first two images, our model generated captions that precisely matched the real ones, while the encoder-decoder model misclassified the first image as a~\emph{resort} and incorrectly detected a~\emph{river}, and in the second image, it mistakenly identified~\emph{green trees}. For the third image, the encoder-decoder model failed to detect both the~\emph{roof of the house} and a~\emph{swimming pool}, whereas our model correctly identified the~\emph{swimming pool}, despite it not being mentioned in the real caption but being present in the image, but it failed to detect the~\emph{roof of the house}. In the supplementary file, we have provided the detailed ablation study. We demonstrated the importance of each component of our model by systematically omitting one component at a time from each version.

\section{Conclusions}
\label{sec:conclusions}
The domain-specific terminologies make remote sensing image captioning a very challenging task. The use of pre-trained embeddings suffer from out-of-vocabulary words. On the other hand, learning word embeddings through the embedding layer of the decoder would limit the model's ability to leverage rich, pre-existing semantic knowledge from the corpus. However, TextGCN generates embeddings by considering the relationships among words at both the sentence level and the corpus level. In this letter, we propose a novel approach that combines TextGCN in a multi-layer encoder-decoder setup. Additionally, we introduce fairness into the traditional and comparison based beam search strategies to generate the final caption. Results show that our approach outperforms various existing RSIC methods.

\bibliographystyle{IEEEtran}
\bibliography{MyPapers}

\begin{thebibliography}{10}
\providecommand{\url}[1]{#1}
\csname url@samestyle\endcsname
\providecommand{\newblock}{\relax}
\providecommand{\bibinfo}[2]{#2}
\providecommand{\BIBentrySTDinterwordspacing}{\spaceskip=0pt\relax}
\providecommand{\BIBentryALTinterwordstretchfactor}{4}
\providecommand{\BIBentryALTinterwordspacing}{\spaceskip=\fontdimen2\font plus
\BIBentryALTinterwordstretchfactor\fontdimen3\font minus \fontdimen4\font\relax}
\providecommand{\BIBforeignlanguage}[2]{{%
\expandafter\ifx\csname l@#1\endcsname\relax
\typeout{** WARNING: IEEEtran.bst: No hyphenation pattern has been}%
\typeout{** loaded for the language `#1'. Using the pattern for}%
\typeout{** the default language instead.}%
\else
\language=\csname l@#1\endcsname
\fi
#2}}
\providecommand{\BIBdecl}{\relax}
\BIBdecl

\bibitem{vinyals2015show}
O.~Vinyals, A.~Toshev, S.~Bengio, and D.~Erhan, ``Show and tell: A neural image caption generator,'' in \emph{Proceedings of the IEEE conference on computer vision and pattern recognition}, 2015, pp. 3156--3164.

\bibitem{karpathy2015deep}
A.~Karpathy and L.~Fei-Fei, ``Deep visual-semantic alignments for generating image descriptions,'' in \emph{Proceedings of the IEEE conference on computer vision and pattern recognition}, 2015, pp. 3128--3137.

\bibitem{hoxha2020toward}
G.~Hoxha, F.~Melgani, and B.~Demir, ``Toward remote sensing image retrieval under a deep image captioning perspective,'' \emph{IEEE Journal of Selected Topics in Applied Earth Observations and Remote Sensing}, vol.~13, pp. 4462--4475, 2020.

\bibitem{yuan2019exploring}
Z.~Yuan, X.~Li, and Q.~Wang, ``Exploring multi-level attention and semantic relationship for remote sensing image captioning,'' \emph{IEEE Access}, vol.~8, pp. 2608--2620, 2019.

\bibitem{huang2020denoising}
W.~Huang, Q.~Wang, and X.~Li, ``Denoising-based multiscale feature fusion for remote sensing image captioning,'' \emph{IEEE Geoscience and Remote Sensing Letters}, vol.~18, no.~3, pp. 436--440, 2020.

\bibitem{wang2020retrieval}
B.~Wang, X.~Zheng, B.~Qu, and X.~Lu, ``Retrieval topic recurrent memory network for remote sensing image captioning,'' \emph{IEEE Journal of Selected Topics in Applied Earth Observations and Remote Sensing}, vol.~13, pp. 256--270, 2020.

\bibitem{li2021recurrent}
Y.~Li, X.~Zhang, J.~Gu, C.~Li, X.~Wang, X.~Tang, and L.~Jiao, ``Recurrent attention and semantic gate for remote sensing image captioning,'' \emph{IEEE Transactions on Geoscience and Remote Sensing}, vol.~60, pp. 1--16, 2021.

\bibitem{hoxha2021novel}
G.~Hoxha and F.~Melgani, ``A novel svm-based decoder for remote sensing image captioning,'' \emph{IEEE Transactions on Geoscience and Remote Sensing}, vol.~60, pp. 1--14, 2021.

\bibitem{zhao2021high}
R.~Zhao, Z.~Shi, and Z.~Zou, ``High-resolution remote sensing image captioning based on structured attention,'' \emph{IEEE Transactions on Geoscience and Remote Sensing}, vol.~60, pp. 1--14, 2021.

\bibitem{zhang2021global}
Z.~Zhang, W.~Zhang, M.~Yan, X.~Gao, K.~Fu, and X.~Sun, ``Global visual feature and linguistic state guided attention for remote sensing image captioning,'' \emph{IEEE Transactions on Geoscience and Remote Sensing}, vol.~60, pp. 1--16, 2021.

\bibitem{wang2022multiscale}
Y.~Wang, W.~Zhang, Z.~Zhang, X.~Gao, and X.~Sun, ``Multiscale multiinteraction network for remote sensing image captioning,'' \emph{IEEE Journal of Selected Topics in Applied Earth Observations and Remote Sensing}, vol.~15, pp. 2154--2165, 2022.

\bibitem{ye2022joint}
X.~Ye, S.~Wang, Y.~Gu, J.~Wang, R.~Wang, B.~Hou, F.~Giunchiglia, and L.~Jiao, ``A joint-training two-stage method for remote sensing image captioning,'' \emph{IEEE Transactions on Geoscience and Remote Sensing}, vol.~60, pp. 1--16, 2022.

\bibitem{liu2022remote}
C.~Liu, R.~Zhao, and Z.~Shi, ``Remote-sensing image captioning based on multilayer aggregated transformer,'' \emph{IEEE Geoscience and Remote Sensing Letters}, vol.~19, pp. 1--5, 2022.

\bibitem{wang2022glcm}
Q.~Wang, W.~Huang, X.~Zhang, and X.~Li, ``Glcm: Global--local captioning model for remote sensing image captioning,'' \emph{IEEE Transactions on Cybernetics}, vol.~53, no.~11, pp. 6910--6922, 2022.

\bibitem{ren2022mask}
Z.~Ren, S.~Gou, Z.~Guo, S.~Mao, and R.~Li, ``A mask-guided transformer network with topic token for remote sensing image captioning,'' \emph{Remote Sensing}, vol.~14, no.~12, p. 2939, 2022.

\bibitem{shen2020remote}
X.~Shen, B.~Liu, Y.~Zhou, J.~Zhao, and M.~Liu, ``Remote sensing image captioning via variational autoencoder and reinforcement learning,'' \emph{Knowledge-Based Systems}, vol. 203, p. 105920, 2020.

\bibitem{hoxha2020new}
G.~Hoxha, F.~Melgani, and J.~Slaghenauffi, ``A new cnn-rnn framework for remote sensing image captioning,'' in \emph{2020 Mediterranean and Middle-East Geoscience and Remote Sensing Symposium (M2GARSS)}.\hskip 1em plus 0.5em minus 0.4em\relax IEEE, 2020, pp. 1--4.

\bibitem{he2016deep}
K.~He, X.~Zhang, S.~Ren, and J.~Sun, ``Deep residual learning for image recognition,'' in \emph{Proceedings of the IEEE conference on computer vision and pattern recognition}, 2016, pp. 770--778.

\bibitem{li2023decoding}
C.~Li, G.~Fang, Y.~Liu, J.~Liu, and L.~Song, ``Decoding silent reading eeg signals using adaptive feature graph convolutional network,'' \emph{IEEE Signal Processing Letters}, 2023.

\bibitem{kim2024stad}
S.~Kim and E.~Park, ``Stad-gcn: Spatial--temporal attention-based dynamic graph convolutional network for retail market price prediction,'' \emph{Expert Systems with Applications}, vol. 255, p. 124553, 2024.

\bibitem{yao2019graph}
L.~Yao, C.~Mao, and Y.~Luo, ``Graph convolutional networks for text classification,'' in \emph{Proceedings of the AAAI conference on artificial intelligence}, vol.~33, no.~01, 2019, pp. 7370--7377.

\bibitem{glorot2010understanding}
X.~Glorot and Y.~Bengio, ``Understanding the difficulty of training deep feedforward neural networks,'' in \emph{Proceedings of the thirteenth international conference on artificial intelligence and statistics}.\hskip 1em plus 0.5em minus 0.4em\relax JMLR Workshop and Conference Proceedings, 2010, pp. 249--256.

\bibitem{das2024unveiling}
S.~Das, A.~Khandelwal, and R.~Sharma, ``Unveiling the power of convolutional neural networks: A comprehensive study on remote sensing image captioning and encoder selection,'' in \emph{2024 International Joint Conference on Neural Networks (IJCNN)}.\hskip 1em plus 0.5em minus 0.4em\relax IEEE, 2024, pp. 1--8.

\bibitem{wang2021cascade}
S.~Wang, Y.~Meng, Y.~Gu, L.~Zhang, X.~Ye, J.~Tian, and L.~Jiao, ``Cascade attention fusion for fine-grained image captioning based on multi-layer lstm,'' in \emph{ICASSP 2021-2021 IEEE International Conference on Acoustics, Speech and Signal Processing (ICASSP)}.\hskip 1em plus 0.5em minus 0.4em\relax IEEE, 2021, pp. 2245--2249.

\bibitem{lu2017exploring}
X.~Lu, B.~Wang, X.~Zheng, and X.~Li, ``Exploring models and data for remote sensing image caption generation,'' \emph{IEEE Transactions on Geoscience and Remote Sensing}, vol.~56, no.~4, pp. 2183--2195, 2017.

\bibitem{qu2016deep}
B.~Qu, X.~Li, D.~Tao, and X.~Lu, ``Deep semantic understanding of high resolution remote sensing image,'' in \emph{2016 International conference on computer, information and telecommunication systems (Cits)}.\hskip 1em plus 0.5em minus 0.4em\relax IEEE, 2016, pp. 1--5.

\bibitem{sharma2017nlgeval}
\BIBentryALTinterwordspacing
S.~Sharma, L.~El~Asri, H.~Schulz, and J.~Zumer, ``Relevance of unsupervised metrics in task-oriented dialogue for evaluating natural language generation,'' \emph{CoRR}, vol. abs/1706.09799, 2017. [Online]. Available: \url{http://arxiv.org/abs/1706.09799}
\BIBentrySTDinterwordspacing

\bibitem{genccay2001pricing}
R.~Gen{\c{c}}ay and M.~Qi, ``Pricing and hedging derivative securities with neural networks: Bayesian regularization, early stopping, and bagging,'' \emph{IEEE transactions on neural networks}, vol.~12, no.~4, pp. 726--734, 2001.

\bibitem{wang2019semantic}
B.~Wang, X.~Lu, X.~Zheng, and X.~Li, ``Semantic descriptions of high-resolution remote sensing images,'' \emph{IEEE Geoscience and Remote Sensing Letters}, vol.~16, no.~8, pp. 1274--1278, 2019.

\bibitem{pennington2014glove}
J.~Pennington, R.~Socher, and C.~D. Manning, ``Glove: Global vectors for word representation,'' in \emph{Proceedings of the 2014 conference on empirical methods in natural language processing (EMNLP)}, 2014, pp. 1532--1543.

\bibitem{mikolov2013efficient}
T.~Mikolov, K.~Chen, G.~Corrado, and J.~Dean, ``Efficient estimation of word representations in vector space,'' \emph{arXiv preprint arXiv:1301.3781}, 2013.

\bibitem{bojanowski2017enriching}
P.~Bojanowski, E.~Grave, A.~Joulin, and T.~Mikolov, ``Enriching word vectors with subword information,'' \emph{Transactions of the association for computational linguistics}, vol.~5, pp. 135--146, 2017.

\end{thebibliography}

\appendix
\section{Modification in Datasets}
While analyzing the original datasets, we identified some errors, which are detailed below:
\subsection{Spelling Errors} We discovered and corrected several misspelled words within the sentences.\\
\textbf{Example: }
\begin{itemize}
    \item\textbf{Dataset:} RSICD~\cite{lu2017exploring}
    \item\textbf{Image:} center\_190.jpg
    \item\textbf{Original Caption:} the polygon center with a cross roof is near a blue~\textcolor{blue}{buliding}
    \item\textbf{Updated Caption:} the polygon center with a cross roof is near a blue~\textcolor{red}{building}
\end{itemize}
\subsection{Variations in Regional Language} We also observed that the words used were not consistent across regions. The final version follows American English conventions.\\
\textbf{Example: }
\begin{itemize}
    \item\textbf{Dataset:} SYDNEY~\cite{qu2016deep}
    \item\textbf{Image:} residential\_22.tif
    \item\textbf{Original Caption:} lots of houses with different~\textcolor{blue}{colours} of roofs arranged densely
    \item\textbf{Updated Caption:} lots of houses with different~\textcolor{red}{colors} of roofs arranged densely 
\end{itemize}
\subsection{Variations in Meaning} We also noticed that different words were used to refer to the same object, leading to confusion. We have standardized this by using a fixed term.\\
\textbf{Example: }
\begin{itemize}
    \item\textbf{Dataset:} UCM~\cite{qu2016deep}
    \item\textbf{Image:} airport\_1.tif
    \item\textbf{Original Caption:} there are some~\textcolor{blue}{planes} and cars in the airport
    \item\textbf{Updated Caption:} there are some~\textcolor{red}{airplanes} and cars in the airport
\end{itemize}
\subsection{Incomplete Sentences or Missing Words} Some sentences were missing one or two words, which we have completed.\\
\textbf{Example: }
\begin{itemize}
    \item\textbf{Dataset:} RSICD~\cite{lu2017exploring}
    \item\textbf{Image:} airport\_106.jpg
    \item\textbf{Original Caption:} is a airport in the middle and many planes in it
    \item\textbf{Updated Caption:}~\textcolor{red}{this} is~\textcolor{blue}{an} airport in the middle and many planes in it
\end{itemize}
In addition to these, we corrected some other minor errors, such as the proper use of~\emph{a} versus~\emph{an}. The updated datasets are provided here (\href{https://drive.google.com/drive/folders/1dBwsMtNWYj7n9osgdsFGUiaEro9RYE8a?usp=drive_link}{click here}).

\begin{figure}[!ht]
  \centering
  \includegraphics[width=0.475\textwidth]{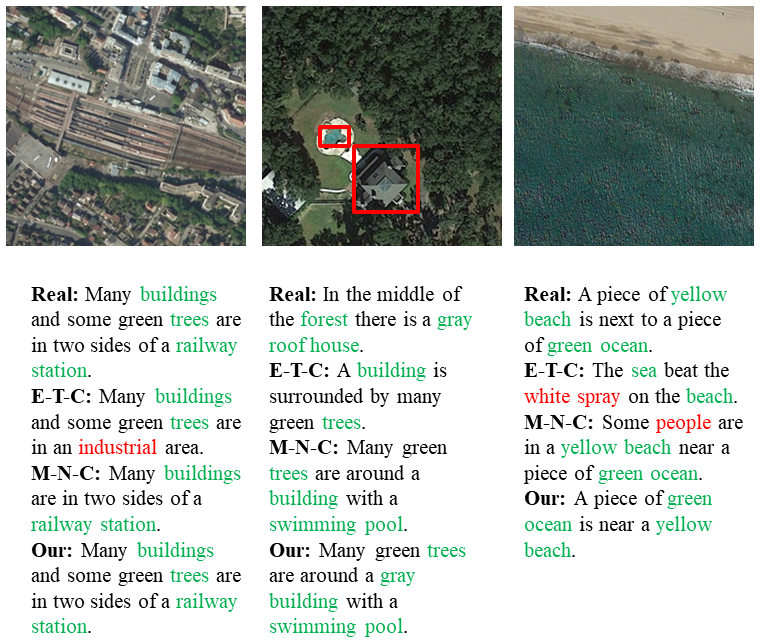}
   \caption{Visual Examples of Proposed RSIC Model}
   \label{fig_visuals}
\end{figure}
\section{Visual Examples}
Figure \ref{fig_visuals} displays test images from the datasets alongside their actual captions and the predicted captions generated by three models: an encoder-decoder model with TextGCN embeddings (E-T-C), our model with trainable embeddings (M-N-C), and our model with TextGCN embeddings (Our). We applied an updated comparison-based beam search technique across all models. The selection of these models aims to demonstrate the effectiveness of each component in our model by systematically omitting one in each variant. In the first image, our model accurately predicts the real caption, whereas E-T-C incorrectly classifies the image as~\emph{industrial area} and M-N-C fails to detect~\emph{trees}. In the second image, our model closely matches the real caption while additionally identifying~\emph{swimming pool} and~\emph{roof's color}, whereas E-T-C missed both~\emph{swimming pool} and~\emph{roof color} and M-N-C missed~\emph{roof color}. It is also evident from the captions that the multi-layer decoder (M-N-C) outperforms the TextGCN embeddings (E-T-C), a finding further supported by the numerical results presented in the main paper.
\end{document}